\documentclass[11pt,letterpaper,twoside,reqno,nosumlimits]{amsart}

\usepackage{etoolbox}
\usepackage{comment}

\patchcmd{\section}{\scshape}{\bfseries}{}{}
\makeatletter
\renewcommand{\@secnumfont}{\bfseries}
\makeatother

\usepackage[dvipsnames]{xcolor}
\patchcmd{\section}{\normalfont}{\normalfont\color{MidnightBlue}}{}{}
\patchcmd{\subsection}{\normalfont}{\normalfont\color{MidnightBlue}}{}{}

\makeatletter
\def\subsubsection{\@startsection{subsubsection}{3}%
\z@{.5\linespacing\@plus.7\linespacing}{-.5em}%
{\normalfont\bfseries}}
\makeatother

\usepackage{fancyhdr}
\usepackage{amsmath,amsfonts,amsbsy,amsgen,amscd,mathrsfs,amssymb,amsthm,mathtools,tensor}

\usepackage{tikz}
\usepackage{overpic}

\usepackage[utf8]{inputenc}
\usepackage{setspace}

\usepackage{graphicx,
    amssymb,
    amsmath,
    amsthm,
    dsfont, 
    xcolor,
    mathtools,
    authblk,
    enumitem,
    tikz,
    mdframed, 
    todonotes,
}

\def\[{\begin{equation} }
\def\]{\end{equation} }
 \usepackage{soul}
\usepackage{subeqnarray}
\usepackage{amsmath, amsfonts} 
\usepackage{algorithm,algorithmic}
\usepackage{graphicx}

\usepackage{tabu}       
\usepackage{siunitx}    
\usepackage{etoolbox}   
\usepackage{booktabs}   
\usepackage{xspace}     
\usepackage{caption}
\usepackage{subcaption}

\newcommand{\RR}{\mathbb{R}}

\newtheorem{definition}{Definition}[section]
\newtheorem{proposition}{Proposition}[section]

\newtheorem{theorem}{Theorem}[section]
\usepackage[all]{xypic}
\usepackage[colorlinks=true,bookmarks=false,citecolor=blue,urlcolor=blue]{hyperref} 
\usepackage{graphicx}
\usepackage{amsmath,amssymb}
\usepackage{environ}
\usepackage[margin=1in]{geometry}
\usepackage{float}

\newcommand{\N}{\mathbb{N}}
\newcommand{\Hc}{\mathcal{H}}
\newcommand{\data}[1]{\textsc{#1}}
\newcommand{\method}{Kernel-IFlow\xspace}

\usepackage[foot]{amsaddr}

\begin{document}

\title[Learning dynamical systems from data, Irregularly-Sampled Time Series]{Learning dynamical systems from data: \\ A simple cross-validation perspective, \\ part III:  Irregularly-Sampled Time Series}

\author{Jonghyeon Lee$^1$}
\address{$^1$ Department of Computing and Mathematical Sciences, Caltech, CA, USA.}
\email{jonghyeonlee98@gmail.com}

\author{ Edward De Brouwer$^2$}
\address{$^2$ ESAT-STADIUS, KU Leuven, Leuven, 3001, Belgium}
\email{ edward.debrouwer@esat.kuleuven.be}
\author{Boumediene Hamzi$^3$}
\address{$^3$ Department of Computing and Mathematical Sciences, Caltech, CA, USA.}
\email{boumediene.hamzi@gmail.com}
\author{Houman Owhadi$^4$}
\address{$^4$Department of Computing and Mathematical Sciences, Caltech, CA, USA. }
\email{ owhadi@caltech.edu}
\maketitle
\begin{abstract}
A simple and interpretable way to learn a dynamical system from data is to interpolate its vector-field with a kernel. In particular, this strategy is highly efficient (both in terms of accuracy and complexity) when the kernel is data-adapted using Kernel Flows (KF)~\cite{Owhadi19} (which uses gradient-based optimization to learn a kernel based on the premise that a kernel is good if there is no significant loss in accuracy if half of the data is used for interpolation). Despite its previous successes, this strategy (based on interpolating the vector field driving the dynamical system) breaks down when the observed time series is not regularly sampled in time. In this work, we propose to address this problem by  approximating a generalization of the flow map of the dynamical system by incorporating time differences between observations in the (KF) data-adapted kernels. We compare our approach with the classical one over different benchmark dynamical systems and show that it significantly improves the forecasting accuracy while remaining simple, fast, and robust.
\end{abstract}

\section{Introduction}
The ubiquity of time series in many domains of science has led to the development of diverse statistical and machine learning forecasting methods. Examples include ARIMA \cite{boxjen70}, GARCH \cite{bollerslev} or LSTM \cite{hochreiter}. Most of these methods require the time series to be regularly sampled in time. Yet, this requirement is not met in many applications. Indeed, irregularly sampled time series commonly arise in  healthcare \cite{jane2016,de2021longitudinal,de2018deep}, finance \cite{spsw} and physics \cite{sbc12} among other fields. 

While adaptations have been proposed, these workarounds tend to consider the irregular sampling issue as a missing values problem, leading to poor performance when the resulting missing rate is very high. Such approaches include (1) the imputation of the missing values (\emph{e.g.} with exponential smoothing \cite{holt57,winters60} or with a Kalman filter \cite{durbinkoopman}), and (2) fast Fourier transforms or Lomb-Scargle periodograms \cite{spsw,vio13}.  This issue has motivated the development of several recent deep learning-based algorithms such as   VS-GRU \cite{vsgru} or Neural-ODE methods \cite{gruodebayes,de2020latent,de2022predicting,rubanova}.

Amongst various learning-based approaches, kernel-based methods hold potential for considerable advantages  in terms of theoretical
analysis, numerical implementation, regularization, guaranteed convergence, automatization, and interpretability \cite{chen2021solving, owhadi2021computational}. Indeed, reproducing kernel Hilbert spaces (RKHS) \cite{CuckerandSmale} have provided strong mathematical foundations for analyzing dynamical systems \cite{bh10, bhcm11,bhcm1,lyap_bh,bh2020a, bh2020b,klus2020data,ALEXANDER2020132520,bh12,bh17} and surrogate modeling (we refer the reader to \cite{santinhaasdonk19} for a survey). Yet, the accuracy of these emulators depends on the kernel, and the problem of selecting a good kernel has received less attention. Recently, the experiments by Hamzi and Owhadi \cite{owhadihamzi} show that when the time series is regularly sampled, Kernel Flows (KF) \cite{Owhadi19} (an RKHS technique) can successfully reconstruct the dynamics of some prototypical chaotic dynamical systems. KFs have subsequently been applied to complex large-scale systems, including climate data \cite{hamzimaulikowhadi,bhkfjpl}. The nonparametric version of KFs has been extended to dynamical systems in \cite{bhkfnp}. A KFs version for  SDEs can be found in  \cite{bhkfsdes}.

Despite its recent successes, we show in this paper that this strategy (based on approximating the vector field of the dynamical system) cannot directly be applied to irregularly sampled time series. Instead, we propose a simple adaptation to the original method that allows to significantly improve forecasting performance when the sampling is irregular. The adaptation is to approximate a generalization of the flow map and can be reduced to adding time delays in between observations to the delay embedding used to feed the method. We demonstrate the benefits of our approach on three prototypical chaotic dynamical systems: the H\'{e}non map, the Van der Pol oscillator, and the Lorenz map. For all, our approach shows significantly improved forecasting accuracy (compared to the original approach).

Specifically, our contributions are as follows:

\begin{itemize}
    \item We show that learning the kernel in kernel ridge regression using our modified approach 
    significantly improves the prediction performance for irregular time series of dynamical systems
    \item Using a delay embedding, we adapt the KF-adapted kernel method algorithm to make multistep predictions
\end{itemize}

The outline of this paper is as follows. In Section 2, we review kernel methods for regularly sampled time series and propose an extension of Kernel Flows to irregularly sampled time series. Section 3 contains a description of our experiments with the Hénon, Van der Pol, and Lorenz systems and a discussion.
The appendix provides a summary of the theory of reproducing kernel Hilbert spaces (RKHS).

\section{Statement of the problem and proposed solution}

\subsection{The problem}
Let $x_{1}, x_{2},...,x_{n}$ be observations from a
deterministic dynamical system in $\mathbb{R}^d$, along with a vector $t=(t_1,\ldots,t_n)$ containing the time of observations. That is, the observation $x_k$ is observed at time $t_k$. Importantly, time differences in between observation $t_{k+1}-t_k$ are not 
necessarily regular. Our goal is to predict 
$x_{n+1}, x_{n+2},\ldots$ given the future sampling times $t_{n+1}, t_{n+2},...$ and the history of the irregularly observed time series ($x_1,...x_{n}$ and $t_1,...,t_n$). 

\subsection{A reminder on kernel methods for regularly sampled time series}\label{secreminder}
The simplest approach to forecasting the time series (employed in \cite{owhadihamzi}) is to assume that $x_1, x_2,\ldots$ is the solution of a discrete dynamical system of the form
\begin{equation}\label{eqjhdbdjehbd}
x_{k+1}=f^\dagger(x_k,\ldots,x_{k-\tau^\dagger+1}),
\end{equation}
with an unknown vector field $f^\dagger$ and time delay $\tau \in \N^*$ (which we will call delay or delay embedding) and
 approximate $f^\dagger$ with a kernel interpolant $f$ of the past data  (a kernel ridge regression model \cite{shawetaylor}) and use the resulting surrogate model $x_{k+1}=f(x_k,\ldots,x_{k-\tau^\dagger+1})$ to predict future state. 

Given $\tau \in \N^*$ (see  \cite{owhadihamzi} for how $\tau$ can be learned in practice), the approximation of the dynamical system can then be recast as that of interpolating $f^\dagger$ from  pointwise measurements
\begin{equation}\label{eqn:fdagger}
f^\dagger(X_k)=Y_k\text{ for }k=1,\ldots, N,
\end{equation}
with $X_k:=(x_k,\ldots,x_{k+\tau-1})$, $Y_k:=x_{k+1}$ and  $N=n-\tau$.
Given a reproducing kernel Hilbert space\footnote{A brief overview of RKHSs is given in the Appendix.} of candidates $\Hc$ for $f^\dagger$, and using the relative error in the RKHS norm $\|\cdot\|_\Hc$ as a loss, the regression of the data $(X_k,Y_k)$ with the kernel $K$ associated with $\Hc$  provides a
minimax optimal approximation \cite{owhadi_scovel_2019}  of  $f^\dagger$ in $ \Hc$.  This regressor (in the presence of measurement noise of variance $\lambda>0$) is
\begin{equation}\label{mean_gp}
f(x)=K(x,X) (K(X,X)+\lambda I)^{-1} Y,
\end{equation}
where  $X=(X_{1},\ldots, X_{N})$, $Y=(Y_{1},\ldots, Y_{N})$,  $K(X,X)$ is the $N\times N $ matrix with entries $K(X_i,X_j)$,  $K(x,X)$ is the $N$ vector with entries $K(x,X_i)$ and $I$ is the identity matrix. This regressor has also a natural interpretation in the setting of Gaussian process (GP) regression: \eqref{mean_gp} is the  conditional mean of the centered GP $\xi\sim \mathcal{N}(0,K)$ with covariance function $K$ conditioned on $\xi(X_k)+\sqrt{\lambda} Z_k=Y_k$ where the $Z_k$ are centered  i.i.d. normal random variables of variance $\lambda$.








\subsection{A reminder on the Kernel Flows (KF) algorithm}\label{subseckf}
The accuracy of any kernel-based method depends on the kernel $K$, and  \cite{owhadihamzi} proposed (in the setting of Subsec.~\ref{secreminder}) to also learn that kernel from the data $(X_k,Y_k)$ with the Kernel Flows (KF)  algorithm \cite{Owhadi19, yooowhadi20, chen2021consistency} which we will now recall.

To describe this algorithm, let  $K_\theta (x,x')$ be a family of kernels parameterized by $\theta$.
Using the notations from Subsection~\ref{secreminder}, the interpolant of the data $(X,Y)$ ($X=(X_1,\ldots,X_N)$ and $Y=(Y_1,\ldots,Y_N)$) obtained with the kernel $K_\theta$ (and a regularization parameter $\lambda >0$ which is chosen as small as possible to ensure accuracy but large enough in case the matrix $K_\theta$ is ill-conditioned) admits the representer formula 

\begin{equation}\label{eqkhgeuyeg}
   u_N(x)= K_\theta(x,X)(K_\theta(X,X)+\lambda I)^{-1}Y
\end{equation}
A fundamental question is then: which $\theta$ should be chosen in \eqref{eqkhgeuyeg}? KF answers that question by learning $\theta$ from data based on the simple premise that a kernel ($K_{\theta}$) is good if the interpolant \eqref{eqkhgeuyeg} does not change much under subsampling of the data. This simple cross-validation concept is then turned into an iterative algorithm as follows.
\\
1. Given $M\leq N$, select a random subset $\{\pi_1,\ldots,\pi_M\}$ of $\{1,\ldots,N\}$ and a random subset $\{\beta_1,\ldots,\beta_{\frac{M}{2}}\}$ of 
$\{\pi_1,\ldots,\pi_M\}$. 
Write $X^\pi$ and $Y^\pi$ for the sub-vectors 
$(X_{\pi_1},\ldots,X_{\pi_M})$ and 
$(Y_{\pi_1},\ldots,Y_{\pi_M})$.
Write $X^\beta$ and $Y^\beta$ for the sub-vectors 
$(X_{\beta_1},\ldots,X_{\beta_\frac{M}{2}})$ and  $(Y_{\beta_1},\ldots,Y_{\beta_\frac{M}{2}})$.
\\
2. Write  $u_{\pi} (x)= K_\theta (x,X^\pi)(K(X^\pi,X^\pi)+\lambda I)^{-1}Y^\pi$
and $u_{\beta} (x)= K_\theta (x,X^\beta)(K(X^\beta,X^\beta)+\lambda I)^{-1}Y^\beta$ for the regressors of  $(X^\pi,Y^\pi)$ and $(X^\beta,Y^\beta)$ obtained with the kernel $K_\theta$.  
\\
3. Write
\begin{equation}
\rho(\theta) :=1-\frac{Y^{\beta,T} (K_\theta (X^\beta,X^\beta)+\lambda I)^{-1}Y^\beta}{Y^{\pi,T} (K_\theta (X^\pi,X^\pi) +\lambda I)^{-1} Y^\pi}\,.
\label{eq:rho}
\end{equation}
Note that (a) when  $\lambda=0$ then $\rho(\theta)$ is the relative square error $\frac{ ||u_\pi -u_{\beta}||^2_{K_\theta}}{||u_\pi||^2_{K_\theta}}$
 between the interpolants $u_\pi$ and $u_\beta$, (b) when 
 $\lambda\geq 0$ then $\rho(\theta)$ is the relative difference $1-\frac{\|u_{\beta}||^2_{K_\theta}+\lambda^{-1}|u_\beta(X^\beta)-Y^\beta|^2}{||u_\pi||^2_{K_\theta}+\lambda^{-1}|u_\pi(X^\pi)-Y^\pi|^2}$ between the regression losses
 (c)  $\rho(\theta)$ lies between 0 and 1 inclusive.
\\
4. Move $\theta$ in the gradient descent direction of $\rho$: $\theta \leftarrow \theta -\eta \nabla_\theta \rho$
\\
5. Repeat until the error reaches a minimum.

\subsection{The problem with irregularly sampled time series}
\label{sec:problem}
The model \eqref{eqjhdbdjehbd} fails to be accurate for irregularly sampled series because it discards the information contained in the $t_k$. 
When the $x_k$ are obtained by sampling a continuous dynamical system, one could consider the following alternative model:
\begin{equation}\label{eqjhdbdjehbd2}
x_{k+1}=x_k+(t_{k+1}-t_k)f^\dagger(x_k)
\end{equation}
While this approach may succeed if the time intervals 
$t_{k+1}-t_k$ are small enough, it will also break down as these time intervals get larger. In our experiments section, we refer to this approach as the \emph{Euler approach}, as it involves learning the Euler discretization of the vector field.  


\subsection{The proposed solution}
To address this issue, we consider the  model
\begin{equation}\label{eqjhdbdjehbd3}
x_{k+1}=f^\dagger(x_k,\Delta_k,\ldots,x_{k-\tau^\dagger+1}, \Delta_{k-\tau^\dagger+1}),
\end{equation}
which incorporates the  time differences $\Delta_k=t_{k+1}-t_{k}$
between observations. That is, we  employ a time-aware time series representations by interleaving observations and time differences. The proposed strategy is then to construct a surrogate model of \eqref{eqjhdbdjehbd3} by regressing  $f^\dagger$ from past data and a kernel $K_\theta$ learned with Kernel Flows as described in Subsec.~\ref{subseckf}. Note that the past data takes the form \eqref{eqn:fdagger} with  $X_k:=(x_k,\Delta_k,\ldots,x_{k+\tau-1}, \Delta_{k+\tau-1})$, $Y_k:=x_{k+1}$ and  $N=n-\tau$.

\section{Experiments}

We conduct numerical experiments on three well-known dynamical systems: the Hénon map, the Van der Pol oscillator, and the Lorenz map. We generate irregularly sampled time series from these dynamical systems using numerical integration and subsequently split the time series into training and test subsets. The time series are subsequently irregularly sampled according to the following scheme. The time interval between each observation $\Delta_k$ is taken to be a multiple of the smallest integration setup used to generate the data $\delta_t$. That is, $\Delta_k = \alpha_k \delta_t$ where $\alpha_k$ is a random integer between $1$ and $\alpha$.  We train the kernel on the training part of the time series and evaluate the forecasting performance of the model. We report both the mean squared error (MSE) and the coefficient of determination ($R^2$).

Given test samples $x_{n+1},x_{n+2},...,x_{N}$ and the predictions $\hat{x}_{n+1}, \hat{x}_{n+2}, ..., \hat{x}_{N}$, the MSE and the coefficient of determination are computed as follows:

\begin{align*}
    MSE &= \frac{1}{N-n}\sum_{i=n+1}^N ||x_i-\hat{x}_i||_2^2 \\
    R^2 &= 1 - \frac{\sum_{i=n+1}^N ||x_i-\hat{x}_i||_2^2}{\sum_{i=n+1}^N ||x_i-  \bar{x}||_2^2}.
\end{align*}

where $\bar{x} = \frac{1}{N-n} \sum_{i=n+1}^N x_i$. The MSE should then be as low as possible and the $R^2$ as high as possible. We note that it is possible to have a negative $R^2$, if the predictor performs worse than the average of the samples.

To showcase the importance of learning the kernel parameters and to include the time difference between subsequent observations, we proceed in three stages. We first report the results of our method when the parameters of the kernel are not learned but rather sampled at random from a uniform ($\mathcal{U}(0,1)$) distribution and when the time delays are not encoded in the input data. In this setup, we distinguish the original KF case and the \emph{Euler} version, as discussed in Subsection~\ref{sec:problem}.
Second, to assess the importance of learning the kernel parameters, we report the model performance when the parameters are learned but the time delays are not encoded in the input data. Lastly, we report the performance of our approach when we both learned the kernel parameters and included the time delays.

For all model variants and dynamical systems, we use the training procedure as described in \cite{owhadihamzi} and used a mini-batch size of 100 temporal observations and minimize $\rho(\theta)$ as in Equation \ref{eq:rho} using stochastic gradient descent. To allow for a notion of uncertainty in the reported metrics, all our experiments use a five-repetition approach where five different kernel initialization are randomly chosen. The minimum of the loss function $\rho(\theta)$ is taken as the value of $\rho(\theta)$ after we run the algorithm for 1000 iterations with a fixed learning rate. The regularization parameter $\lambda$ has been set to $10^{-5}$ for all experiments.

In all of our examples, we used a kernel with 24 parameters that is a linear combination of the triangular, Gaussian, Laplacian, locally periodic kernels, and the quadratic kernel. This kernel was originally designed to account for the periodic, quasi-triangular, and inverted quadratic behavior of the Hénon map, but training this kernel with irregular Kernel Flows gives accurate results for the Van der Pol and Lorenz systems as well.

\begin{multline}
    K_{\theta} (x,y) = a^2_1 e^{-\frac{||x-y||^2}{2a^2_2}}+\beta^2_1(x^T y + \beta_2)^2 + c^2_1 (c^2_2+ c^2_3 ||x-y||)^{-\frac{1}{2}}+\gamma^2_1 (\gamma^2_2+ ||x-y||^2)^{-\gamma_{3}}+d^2_{1} (1+d^{-2}_{2}
     ||x-y||)^{-1}
     \\
     + p^2_{1} p_{2} \max(0,1-p_{3})+p_{4} e^{-\frac{||x-y||^2}{2p^2_{5}}}+q^2_{1} e^{\frac{-\sin^2 (\pi q^{-1}_{2}||x-y||^2)}{q^2_{3}}} e^{-\frac{||x-y||^2}{q^2_{4}}}+ s^2_{1} e^{-\frac{\sin(\pi s^{-1}_{2} ||x-y||^2)}{s^2_{3}}},
     \end{multline}

where $\theta = (a_1,a_2,\beta_1,\beta_2,c_1,c_2,c_3,\gamma_1,\gamma_2,\gamma_3,d_1,d_2,p_1,p_2,p_3,p_4,p_5,q_1,q_2,q_3,q_4,s_1,s_2,s_3).$

\paragraph{\bf Multi-step predictions:}

By learning the dynamical systems of interest, we aim to deliver accurate forecasting predictions over the longest horizon possible. However, due to the chaotic nature of the studied dynamical systems, this horizon is intrinsically limited. To use most of the testing section of the time series, we then predict the future of the time series in chunks. That is, for a horizon $h$ and for a delay embedding with delay $d$, we split the test time series in chunks of lengths $h+d$. For each of these chunks, we use the $d$ first samples as input to our model and predict over the $h$ remaining samples in the chunk. We eventually aggregate the predictions of all samples overall chunks together to compute the reporting metrics.

\paragraph{\bf Overview.}
Recapping, we will compare 5 approaches:
\renewcommand{\theenumi}{\Alph{enumi}}
\begin{enumerate}
\item Regressing model  \eqref{eqjhdbdjehbd3} with a kernel learnt using KF  (which we call irregular KF or Kernel-IFlow).
\item 
Regressing model \eqref{eqjhdbdjehbd}  with a  kernel learnt using KF (which we call regular KF).
\item Regressing model \eqref{eqjhdbdjehbd2} with a kernel learnt using KF (which we call the Euler version).
\item Regressing model  \eqref{eqjhdbdjehbd3} without learning the kernel.
\item Regressing model \eqref{eqjhdbdjehbd} without learning the kernel.

\end{enumerate}

Table \ref{tab:results} summarizes results obtained in the following sections.

\begin{table}[tbp]
  \caption{%
    Test performance of the different datasets. We report the means along with standard deviations of the mean squared error (MSE) and coefficient of determination ($R^2$) on the forecasting task. As Hénon is not a time-continuous map, the Euler version of KF is not applicable in this case. For readability, we abstain from reporting the exact numbers when MSE is larger than one and $R^2$ larger lower than 0.
  }
  \label{tab:results}
  \vskip 0.1in
  \centering
  \small
  \newcommand{\NA}{---}
  \sisetup{
    detect-all           = true,
    table-format         = 2.1(2),
    separate-uncertainty = true,
    mode                 = math,
    table-text-alignment = center,
    tight-spacing,
  }
  \robustify\bfseries
  \renewrobustcmd{\bfseries}{\fontseries{b}\selectfont}
  \renewrobustcmd{\boldmath}{}
  \let\b\bfseries
  \setlength{\tabcolsep}{3.0pt}
  \begin{tabu}{lcccccc}
    \toprule
    \textsc{Method}      & \multicolumn{2}{c}{{\scriptsize\data{Hénon}}}
                         & \multicolumn{2}{c}{{\scriptsize\data{Lorenz}}} & \multicolumn{2}{c}{{\scriptsize\data{Van der Pol}}} \\
    \midrule
     & \textsc{MSE} & \textsc{$R^2$} & \textsc{MSE} & \textsc{$R^2$} & \textsc{MSE} & \textsc{$R^2$} \\
    \midrule
     (A) \method  &  \num{0.024 \pm 0.015} & \num{0.869 \pm 0.081} & \num{0.003 \pm 0.003} & \num{0.967 \pm 0.029} & \num{0.001 \pm 0.001} & \num{0.998 \pm 0.002}\\
          (B) KernelFlow         &  \num{0.190 \pm 0.008}         & \num{-0.050 \pm 0.041}  & \num{0.026 \pm 0.015} & \num{0.700 \pm 0.170} & \textsc{$>>1$} & \textsc{$<<0$}
\\
    (C) KernelFlow (\emph{Euler})           &  /         &  /  & \num{0.005 \pm 0.002} & \num{0.947 \pm 0.023} & \textsc{$>>1$} & \textsc{$<<0$}\\
     \midrule
     (D) - no learning  &  \textsc{$>>1$} & \textsc{$<<0$} & \textsc{$>>1$} & \textsc{$<<0$} & \textsc{$>>1$} & \textsc{$<<0$}\\
     (E) - no learning &  \textsc{$>>1$} & \textsc{$<<0$} & \textsc{$>>1$} & \textsc{$<<0$} & \textsc{$>>1$} & \textsc{$<<0$}\\
    \bottomrule
  \end{tabu}
\end{table}

\subsection{Hénon map}

\begin{figure}[!htb]
   \begin{subfigure}[t]{0.48\textwidth}
     \centering
     \includegraphics[width=0.95\columnwidth]{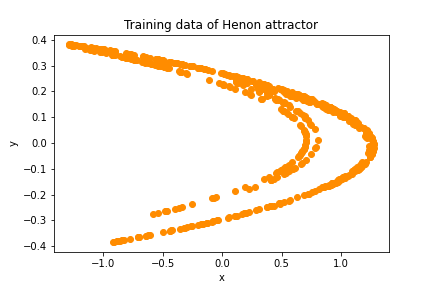}
     \caption{Graph of Hénon attractor from irregularly sampled time series}\label{fig:henon_irregular_attractor}
   \end{subfigure}\hfill
   \begin{subfigure}[t]{0.48\textwidth}
     \centering
     \includegraphics[width=0.95\columnwidth]{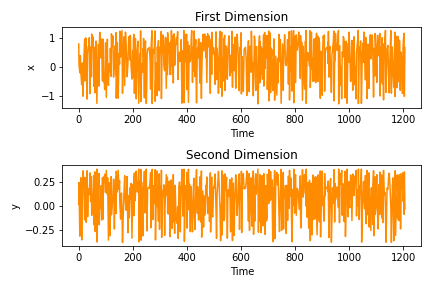}
     \caption{Graph of irregularly sampled time series for Hénon map}\label{Fig:henon_irregular_ts}
   \end{subfigure}
   \caption{Irregularly sampled training data for Hénon map}
   \label{fig:henon_attractor_irregular}
\end{figure}

\begin{figure}[!htb]
   \begin{subfigure}[t]{0.48\textwidth}
     \centering
     \includegraphics[width=0.95\columnwidth]{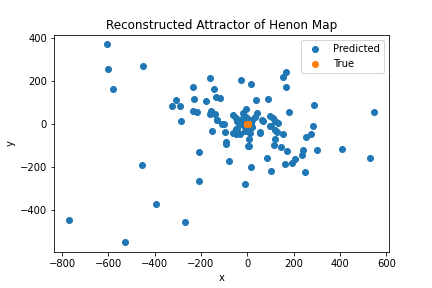}
     \caption{Approach (E). Attractor Reconstruction by regressing model \eqref{eqjhdbdjehbd} without learning the kernel (horizon has been reduced to 1).}\label{fig:henon_no_learning_attractor}
   \end{subfigure}\hfill
   \begin{subfigure}[t]{0.48\textwidth}
     \centering
     \includegraphics[width=0.95\columnwidth]{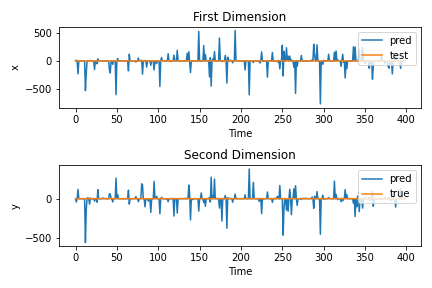}
     \caption{Approach (E). Time series reconstruction by regressing model \eqref{eqjhdbdjehbd} without learning the kernel (horizon has been reduced to 1).}
     \label{fig:henon_ts_no_learning}
   \end{subfigure}
   \caption{Hénon map reconstructions when the kernel parameters are not learnt (regression of model \eqref{eqjhdbdjehbd} without learning the kernel).}
   \label{fig:henon_attractor_no_learning}
\end{figure}

\begin{figure}[!htb]
\begin{subfigure}[t]{0.48\textwidth}
     \centering
     \includegraphics[width=.95\linewidth]{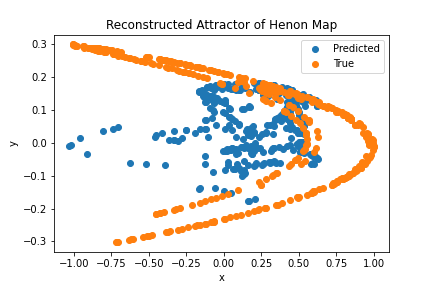}
     \caption{Approach (B). With regular Kernel Flows (regression of model \eqref{eqjhdbdjehbd} with a  kernel learnt using KF)}\label{Fig:Data1}
   \end{subfigure}\hfill
  \begin{subfigure}[t]{0.48\textwidth}
     \centering
     \includegraphics[width=.95\linewidth]{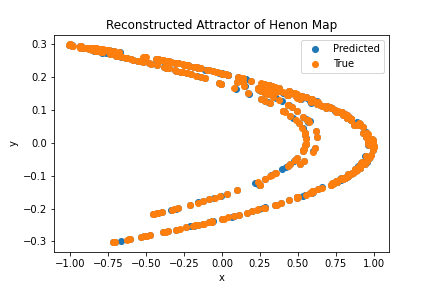}
     \caption{Approach (A). With irregular Kernel Flows (regression of model \eqref{eqjhdbdjehbd3} with a kernel learnt using KF)}\label{Fig:HenonData4}
    \end{subfigure}
  \caption{Hénon map attractor reconstructions with learnt kernels.}
   \label{fig:aandb}
\end{figure}

\begin{figure}[!htb]
   \begin{subfigure}[t]{0.48\textwidth}
     \centering
     \includegraphics[width=.95\linewidth]{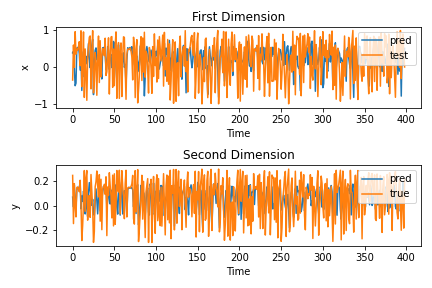}
      \caption{Approach (B). With regular Kernel Flows}\label{Fig:HenonData5}
   \end{subfigure}\hfill
   \begin{subfigure}[t]{0.48\textwidth}
     \centering
     \includegraphics[width=.95\linewidth]{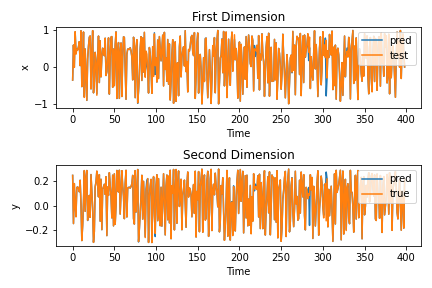}
     \caption{Approach (A). With irregular Kernel Flows}\label{Fig:HenonData6}
   \end{subfigure}
   \caption{Reconstruction (prediction) of the test time series of the Hénon 
 map.}
\end{figure}

Consider the Hénon map with $a=1.4,b=0.3$

\begin{equation}
    x_{n+1}= 1- a x_n^2+y_n ,     y_{n+1}= b x_n
\end{equation}
\newline
 We have repeated our experiments five times with a delay embedding of 1, a learning rate $\eta$ of 0.1, a prediction horizon $h$ of 5, a maximum time difference $\alpha$ of 3, and have trained the model on 600 irregularly sampled points chosen between $T=0$ to $T=1199$ to predict the next 400 points. The panels in Figure \ref{fig:henon_attractor_irregular} display the attractor of the H\'enon map and the irregularly sampled time series from which it was formed.
Figure \ref{fig:henon_attractor_no_learning} shows that approach (E) cannot reconstruct the attractor or make an accurate forecast of the time series because it makes no attempt at learning the kernel and ignores time differences in the training data. Figures \ref{Fig:HenonData4} and \ref{Fig:HenonData6} illustrate that embedding the time delay in the kernel (approach (A))  significantly improves the reconstruction of the attractor of the H\'enon map and the prediction of the time series. Table \ref{tab:results} displays the forecasting performance of the different methods. We observe that if the kernel is not learned (if the kernel is not data adapted), then the underlying method is unable to learn an accurate representation of the dynamical system. However, if the parameters of the kernel are learned, then our proposed approach (A) clearly outperforms the regular KF (approach (B) in \ref{Fig:Data1} and \ref{Fig:HenonData5}). As for the Euler version, it is not applicable in this case as Hénon is not a continuous map. We have also included in Table \ref{tab:alphahenon} in the appendix the results of an experiment that varies $\alpha$ for the irregular and regular KF algorithms.

\renewcommand{\thesubfigure}{\roman{subfigure}}

\renewcommand{\thesubfigure}{\roman{subfigure}}

\subsection{Van der Pol oscillator}

The second dynamical system of interest is the Van der Pol oscillator represented by

\begin{equation}
    \frac{dx}{dt}=\frac{1}{\epsilon} f(x,y)\  ,\
    \frac{dy}{dt}= g(x,y)
\end{equation}
\newline
where $f(x,y)=y-\frac{27}{4} x^2 (x+1),\ g=-\frac{1}{2}-x\  ,\epsilon=0.01$.
\newline

\begin{figure}[!htb]
   \begin{subfigure}[t]{0.48\textwidth}
     \centering
     \includegraphics[width=.95\linewidth]{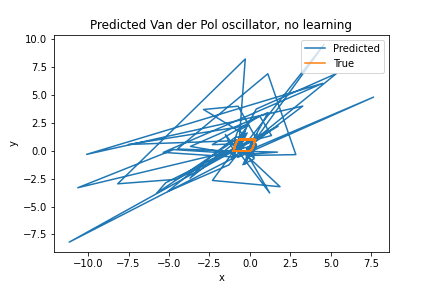}
     \caption{Attractor Reconstruction.}\label{Fig:VDPData1}
   \end{subfigure}\hfill
   \begin{subfigure}[t]{0.48\textwidth}
     \centering
     \includegraphics[width=.95\linewidth]{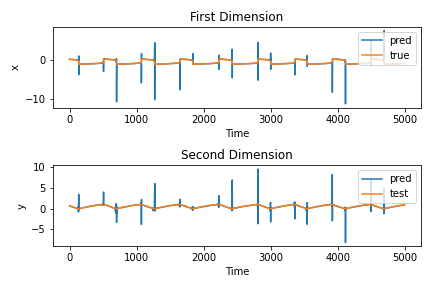}
     \caption{Time series reconstruction.}\label{Fig:VDPData2}
   \end{subfigure}
   \caption{Van der Pol oscillator without learning the kernel (horizon has been reduced to 1).}\label{vdp1}
\end{figure}

\begin{figure}[!htb]
   \begin{subfigure}[t]{0.48\textwidth}
     \centering
     \includegraphics[width=.95\linewidth]{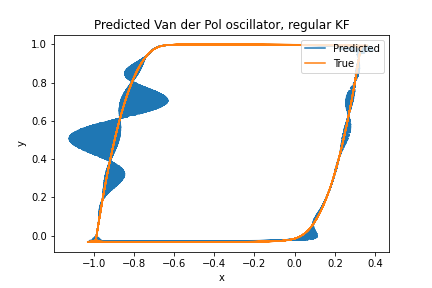}
     \caption{Approach (B): with regular Kernel Flows (the horizon has been reduced to 4).}\label{Fig:VDPData3}
    \end{subfigure}\hfill
   \begin{subfigure}[t]{0.48\textwidth}
     \centering
     \includegraphics[width=.95\linewidth]{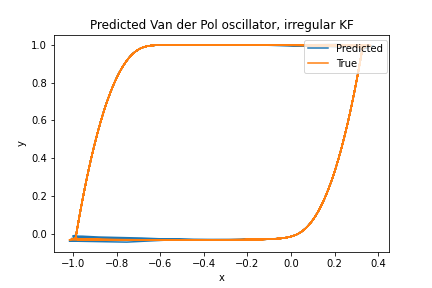}
     \caption{Approach (A): with irregular Kernel Flows}\label{Fig:VDPData4}
    \end{subfigure}
\caption{Van der Pol attractor reconstruction.}\label{vdp2}
\end{figure}

\begin{figure}[!htb]
   \begin{subfigure}[t]{0.48\textwidth}
     \centering
     \includegraphics[width=.95\linewidth]{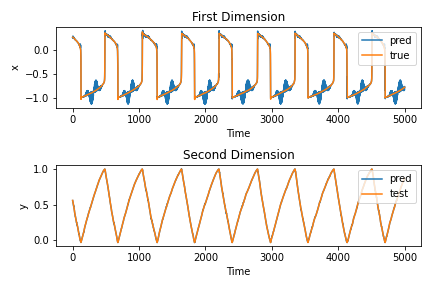}
     \caption{Approach (B): with regular Kernel Flows}\label{Fig:VDPData5}
   \end{subfigure}\hfill
   \begin{subfigure}[t]{0.48\textwidth}
     \centering
     \includegraphics[width=.95\linewidth]{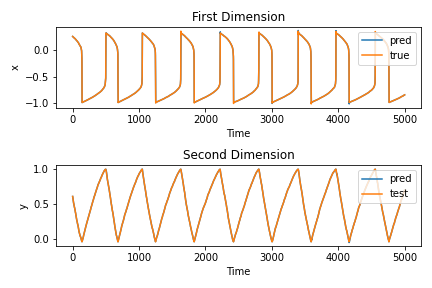}
     \caption{Approach (A): With irregular Kernel Flows}\label{Fig:VDPData6}
   \end{subfigure}
   \caption{Reconstruction of the test time series of the Van der Pol oscillator with irregular and regular Kernel Flows.}\label{vdp3}
\end{figure}

Here, we have used a prediction horizon $h$ of 10, a learning rate $\eta$ of 0.01, a maximum time difference $\alpha$ of 5, and a delay embedding of 1. As evident from Table \ref{tab:results} and Figure \ref{vdp3}, our proposed approach (A) is the only one able to extract any meaningful representation of the dynamical system (for example, correctly predicting the behavior immediately after the "jumps" in the top graph of Figure \ref{Fig:VDPData6}). Figure \ref{vdp1} shows that approach (E) fails to reconstruct the time series as it neither learns the kernel nor considers the time differences between the observations. In Figures \ref{Fig:VDPData3} and \ref{Fig:VDPData5}, which illustrate approach (B), the learned kernel fails to accurately predict the behavior after the "jumps" as the time differences between the observations are disregarded. In the appendix, we also tested the KF algorithms for greater values of $\alpha$; the results in Table \ref{tab:alphavdp} show that irregular KF outperforms the regular and Euler versions for $\alpha=6$ and 7.

\subsection{Lorenz}

Our third example  is the Lorenz system described by the following system of differential equations:

\begin{equation}
    \dot{x} = \sigma (y-x)  , \ \dot{y} = x(\rho-z) - y  , \ \dot{z} = xy - \beta z\,,
\end{equation}
\newline
with standard parameter values $\sigma = 10 , \ \rho = 28  
\ ,\beta = \frac{8}{3}$.

\begin{figure}[!htb]
   \begin{subfigure}[t]{0.48\textwidth}
     \centering
     \includegraphics[width=0.95\linewidth]{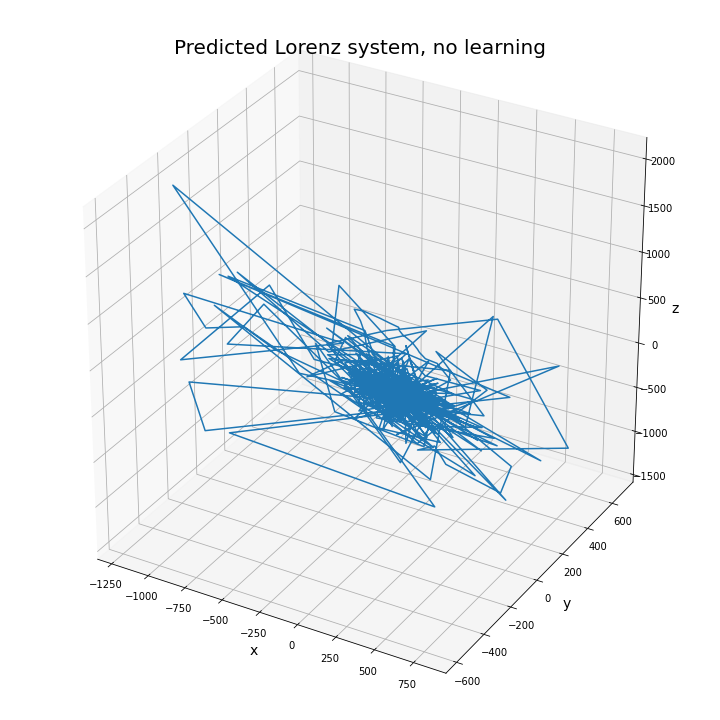}
     \caption{Attractor reconstruction.}\label{Fig:LorenzData1}
   \end{subfigure}\hfill
   \begin{subfigure}[t]{0.48\textwidth}
     \centering
     \includegraphics[width=.95\linewidth]{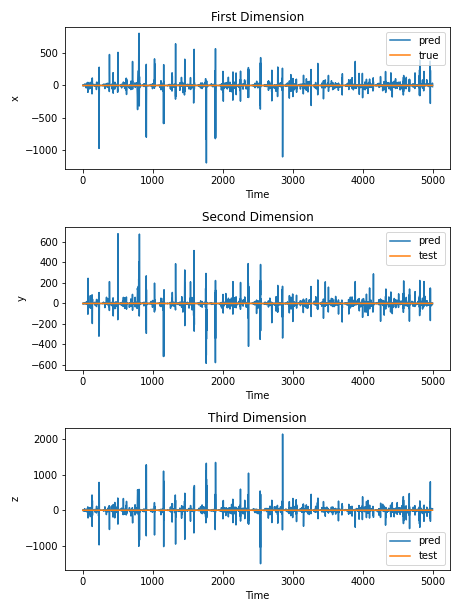}
     \caption{Time series reconstruction. }\label{Fig:LorenzData2}
   \end{subfigure}
   \caption{Lorenz map without learning the kernel (horizon has been reduced to 1).}\label{figlorat}
\end{figure}

\begin{figure}[!htb]
   \begin{subfigure}[t]{0.48\textwidth}
     \centering
     \includegraphics[width=.95\linewidth]{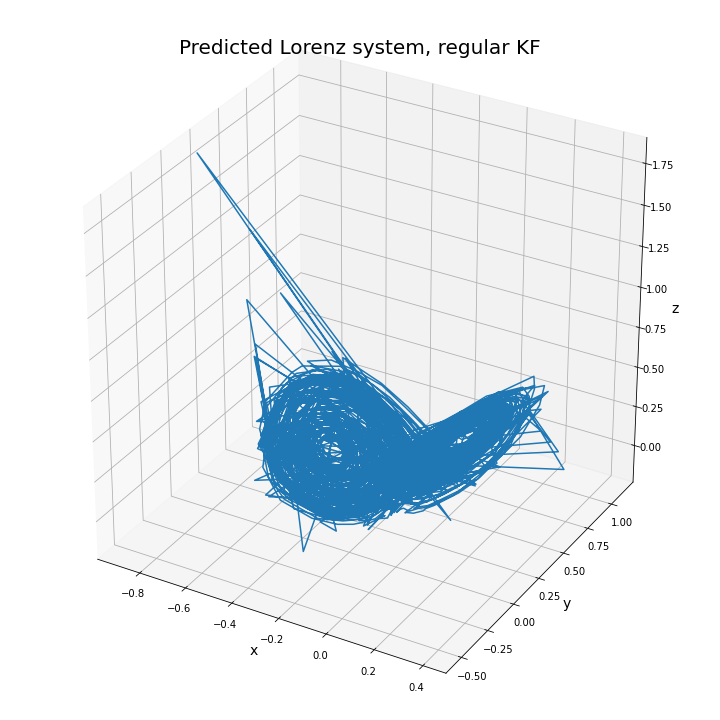}
     \caption{Approach (B): With regular Kernel Flows}\label{Fig:LorenzData3}
   \end{subfigure}\hfill
   \begin{subfigure}[t]{0.48\textwidth}
     \centering
     \includegraphics[width=.95\linewidth]{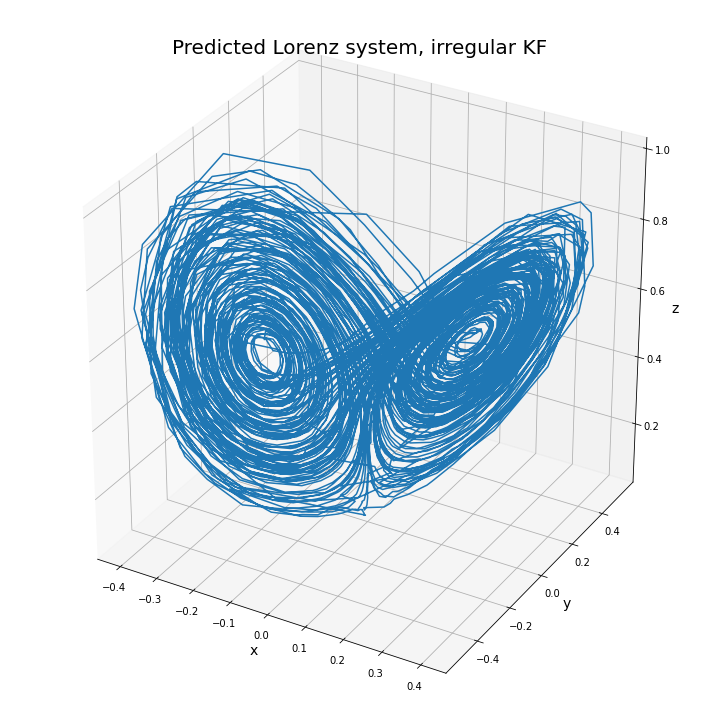}
     \caption{Approach (A): with irregular Kernel Flows}\label{Fig:LorenzData4}
   \end{subfigure}
  \caption{Lorenz map attractor reconstruction with learnt kernel.}\label{figlorat2}
\end{figure}

\begin{figure}[!htb]
   \begin{subfigure}[t]{0.48\textwidth}
     \centering
     \includegraphics[width=.95\linewidth]{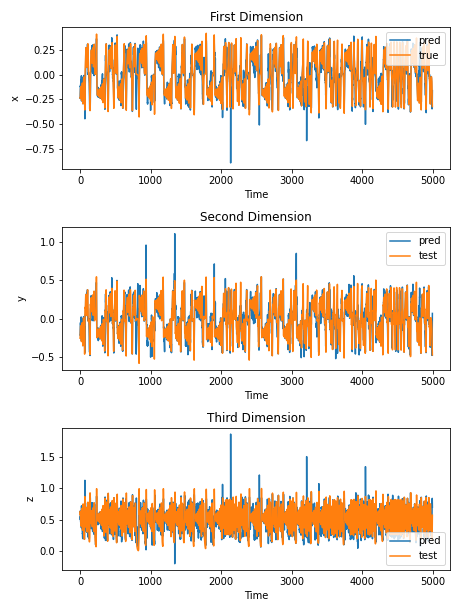}
     \caption{Approach (B): with regular Kernel Flows}\label{Fig:LorenzData5}
   \end{subfigure}\hfill
   \begin{subfigure}[t]{0.48\textwidth}
     \centering
     \includegraphics[width=.95\linewidth]{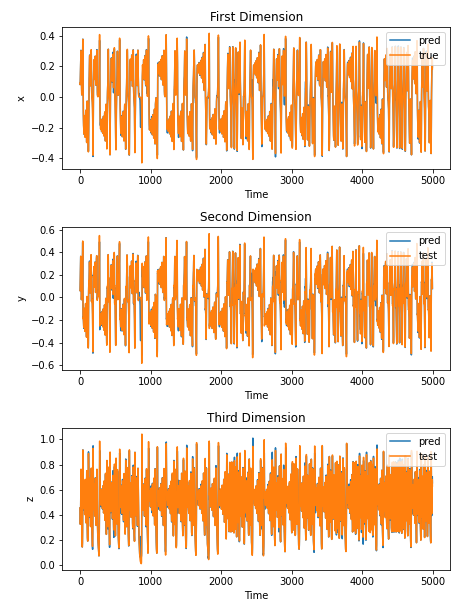}
     \caption{Approach (A): with irregular kernel flows}\label{Fig:LorenzData6}
   \end{subfigure}
   \caption{Reconstruction of the test time series of the Lorenz map with irregular and regular Kernel Flows.}\label{figlorat3}
\end{figure}

Our parameters include a delay embedding  of 2, a learning rate $\eta = 0.01$, a prediction horizon $h=20$, a maximum time difference $\alpha = 5$, 5000 points used for training and the 5000 for testing. Figures \ref{figlorat}, \ref{Fig:LorenzData3} and \ref{Fig:LorenzData5} show that not learning the kernel or not including time differences leads to poor reconstructions of the attractor of the Lorenz system even if the time horizon is 1,  compared to the results of approach (A) shown in Figures \ref{Fig:LorenzData4} and \ref{Fig:LorenzData6}. However, as observed in Table \ref{tab:results}, the Euler version of KF leads to satisfying results, close to (but not as good as) the ones obtained with our proposed approach (A). In the appendix, we have also included additional experiments in which we have re-run the KF algorithms with different values of $\alpha$, $\eta$, $h$, and training epochs.

\paragraph{\bf Remark: }
To include new measurements when approximating the dynamics from data without repeating the learning process, see \cite{bhkfnp}. This can be done by working in Newton basis as in \cite{PAZOUKI2011575} (see also section 4 of \cite{santinhaasdonk19}).
The Newton basis is just another basis for the space spanned by the kernel on the points, i.e., $\mbox{span}\{K(., x_1), ...., K(., x_N)\} = \mbox{span}\{v_1, ..., v_N\}$.

The kernel expansion of $f$ writes as $f(x)=\sum_{i=1}^N c_i K(x,x_i) = \sum_{i=1}^N b_i v_i(x)$ with $<v_i, v_j>_H = \delta_{ij}$ (i.e., the basis is orthonormal in the RKHS inner product).

If we add a new point $x_{N+1}, ..., x_{N+m}$, we'll have corresponding elements $v_{N+1}, ..., v_{N+m}$ of the Newton basis, still orthonormal to the previous ones. So we will have a new interpolant $f_{\mbox{new}}(x)=\sum_{i=1}^{N+m}  b_i v_i(x)$ that can be rewritten in terms of the old interpolant as

$$ f_{\mbox{new}}(x) = \sum_{i=1}^{N+m} c_i v_i(x) = f(x) + \sum_{i=N+1}^{N+m} c_i v_i(x),$$

where $f$ can still be written in terms of the basis K, but with different coefficients $c'$.

If $A$ is the kernel matrix on the first $N$ points, on can compute a Cholesky factorization $A = L L^T$ with $L$ lower triangular. Let $B:=L^{-T}$, then $v_j(x) = \sum_{i=1}^N (B)_{ij} K(x, x_i)$.

When we add new points, we have an updated kernel matrix $A'$, and the Cholesky factor of $A$ can be easily updated to the one of $A'$.

\section{Conclusion}

Our numerical experiments demonstrate that embedding the time differences between the observations in the kernel considerably improves the forecasting accuracy with irregular time series. Though we have focused on a few examples, the success of our proposed approach (A) has raised the question of whether it can be extended to other systems, including those described by partial and stochastic differential equations, as well as complex real-world data.

\section{Appendix}

\subsection{Reproducing Kernel Hilbert Spaces (RKHS)}
\ 
We give a brief overview of reproducing kernel Hilbert spaces as used in statistical learning
theory ~\cite{CuckerandSmale}. Early work developing
the theory of RKHS was undertaken by N. Aronszajn~\cite{aronszajn50reproducing}.

\begin{definition} Let  ${\mathcal H}$  be a Hilbert space of functions on a set ${\mathcal X}$.
Denote by $\langle f, g \rangle$ the inner product on ${\mathcal H}$   and let $\|f\|= \langle f, f \rangle^{1/2}$
be the norm in ${\mathcal H}$, for $f$ and $g \in {\mathcal H}$. We say that ${\mathcal H}$ is a reproducing kernel
Hilbert space (RKHS) if there exists a function $K:{\mathcal X} \times {\mathcal X} \rightarrow \RR$
such that\\
 i. $K_x:=K(x,\cdot)\in {\mathcal{H}}$ for all $x\in {\mathcal{X}}$.\\
ii. $K$ spans ${\mathcal H}$: ${\mathcal H}=\overline{\mbox{span}\{K_x~|~x \in {\mathcal X}\}}$.\\
 iii. $K$ has the {\em reproducing property}:
$\forall f \in {\mathcal H}$, $f(x)=\langle f,K_x \rangle$.\\
$K$ will be called a reproducing kernel of ${\mathcal H}$. ${\mathcal H}_K$  will denote the RKHS ${\mathcal H}$
with reproducing kernel  $K$ where it is convenient to explicitly note this dependence.
\end{definition}

The important properties of reproducing kernels are summarized in the following proposition.
\begin{proposition}\label{prop1} If $K$ is a reproducing kernel of a Hilbert space ${\mathcal H}$, then\\
i. $K(x,y)$ is unique.\\
ii.  $\forall x,y \in {\mathcal X}$, $K(x,y)=K(y,x)$ (symmetry).\\
iii. $\sum_{i,j=1}^q\beta_i\beta_jK(x_i,x_j) \ge 0$ for $\beta_i \in \RR$, $x_i \in {\mathcal X}$ and $q\in\mathbb{N}_+$
(positive definiteness).\\
iv. $\langle K(x,\cdot),K(y,\cdot) \rangle=K(x,y)$.
\end{proposition}
Common examples of reproducing kernels defined on a compact domain $\mathcal{X} \subset \mathrm{R}^n$ are the 
(1) constant kernel: $K(x,y)= k > 0$
(2) linear kernel: $K(x,y)=x\cdot y$
(3) polynomial kernel: $K(x,y)=(1+x\cdot y)^d$ for $d \in \N_+$
(4) Laplace kernel: $K(x,y)=e^{-||x-y||_2/\sigma^2}$, with $\sigma >0$
(5)  Gaussian kernel: $K(x,y)=e^{-||x-y||^2_2/\sigma^2}$, with $\sigma >0$
(6) triangular kernel: $K(x,y)=\max \{0,1-\frac{||x-y||_2^2}{\sigma} \}$, with $\sigma >0$.
(7) locally periodic kernel: $K(x,y)=\sigma^2 e^{-2 \frac{ \sin^2(\pi ||x-y||_2/p)}{\ell^2}}e^{-\frac{||x-y||_2^2}{2 \ell^2}}$, with $\sigma, \ell, p >0$.

\begin{theorem} \label{thm1}
Let $K:{\mathcal X} \times {\mathcal X} \rightarrow \RR$ be a symmetric and positive definite function. Then there
exists a Hilbert space of functions ${\mathcal H}$ defined on ${\mathcal X}$   admitting $K$ as a reproducing Kernel.
Conversely, let  ${\mathcal H}$ be a Hilbert space of functions $f: {\mathcal X} \rightarrow \RR$ satisfying
$\forall x \in {\mathcal X}, \exists \kappa_x>0,$ such that $|f(x)| \le \kappa_x \|f\|_{\mathcal H},
\quad \forall f \in {\mathcal H}. $
Then ${\mathcal H}$ has a reproducing kernel $K$.
\end{theorem}


\begin{theorem}\label{thm4}
 Let $K(x,y)$ be a positive definite kernel on a compact domain or a manifold $X$. Then there exists a Hilbert
space $\mathcal{F}$  and a function $\Phi: X \rightarrow \mathcal{F}$ such that
$$K(x,y)= \langle \Phi(x), \Phi(y) \rangle_{\mathcal{F}} \quad \mbox{for} \quad x,y \in X.$$
 $\Phi$ is called a feature map, and $\mathcal{F}$ a feature space\footnote{The dimension of the feature space can be infinite, for example in the case of the Gaussian kernel.}.
\end{theorem}

\subsection{Function Approximation in RKHSs: An Optimal Recovery Viewpoint} 
In this section, we review function approximation in RKHSs from the point of view of optimal recovery as discussed in \cite{owhadi_scovel_2019}. 

\paragraph{Problem {\bf P}:} Given input/output data $(x_1, y_1),\cdots , (x_N , y_N ) \in \mathcal{X} \times \mathbb{R}$,  recover an unknown function $u^{\ast}$ mapping $\mathcal{X}$ to $\mathbb{R}$ such that
$u^{\ast}(x_i)=y_i$ for $i \in \{1,...,N\}$.

In the setting of optimal recovery, \cite{owhadi_scovel_2019}  Problem {\bf P} can be turned into a well-posed problem by restricting candidates for $u$ to belong to a Banach space of functions $\mathcal{B}$ endowed with a norm $||\cdot||$ and identifying the optimal recovery as the minimizer of the relative error

\begin{equation} \label{game}
    \mbox{min}_v\mbox{max}_u \frac{||u-v||^2}{||u||^2}, 
\end{equation} 
where the max is taken over $u \in \mathcal{B}$ and the min is taken over candidates in $v \in \mathcal{B}$ such that $v(x_i)=u(x_i)=y_i$. For the validity of the constraints $u(x_i) = y_i$,  $\mathcal{B}^{\ast}$, the dual space of $\mathcal{B}$, must contain delta Dirac functions $\phi_i(\cdot)=\delta(\cdot-x_i)$. This problem can be stated as a game between Players I and II and can then be represented as
  \begin{equation}\label{eqdkjdhkjhffORgameban}
\text{\xymatrixcolsep{0pc}\xymatrix{
\text{(Player I)} & u\ar[dr]_{\max}\in \mathcal{B}    &      &v\ar[ld]^{\min}\in L(\Phi,\mathcal{B}) &\text{(Player II)}\\
&&\frac{\|u-v(u)\|}{\|u\|}\,.& &
}}\,
\end{equation}

If $||\cdot||$ is quadratic, i.e. $||u||^2=[Q^{-1}u,u] $ where $[\phi, u]$ stands for the duality product between $\phi \in \mathcal{B}^{\ast}$ and $u \in \mathcal{B}$ and $Q : \mathcal{B}^{\ast}\rightarrow \mathcal{B}$ is a positive symmetric linear bijection (i.e. such that $[\phi, Q \phi] \ge  0$ and $[\psi, Q \phi ] = [\phi, Q \psi]$ for $\phi,\psi \in \mathcal{B}^{\ast} $). In that case the optimal solution of (\ref{game}) has the explicit form 
\begin{equation}\label{sol_rep}v^{\ast}=\sum_{i,j=1}^{N}u(x_i) A_{i,j} Q \phi_j, \end{equation}
where   $A=\Theta^{-1}$ and $\Theta \in \RR^{N \times N}$ is a Gram matrix with entries $\Theta_{i,j}=[\phi_i,Q\phi_j]$.

To recover the classical representer theorem, one defines the reproducing kernel $K$ as $$K(x,y)=[\delta(\cdot-x),Q\delta(\cdot-y)]$$ 
In this case, $(\mathcal{B},||\cdot ||)$ can be seen as an RKHS endowed with the norm
$$||u||^2=\mbox{sup}_{\phi \in \mathcal{B}^\ast}\frac{(\int \phi(x) u(x) dx)^2}{(\int \phi(x) K(x,y) \phi(y) dx dy)}$$
and (\ref{sol_rep}) corresponds to the classical representer theorem 
\begin{equation}\label{eqkjelkjefffhb}
v^{\ast}(\cdot) = y^T AK(x,\cdot),
\end{equation} 
 using the vectorial notation $y^T AK(x,\cdot)=\sum_{i,j=1}^{N}y_iA_{i,j}K(x_j,\cdot)$ with $y_i=u(x_i)$, $A=\Theta^{-1}$ and $\Theta_{i,j} =K(x_i,x_j)$.
  
 Now, let us consider the problem of learning the kernel from data. As introduced in \cite{Owhadi19}, the method of KFs is based on the premise that \emph{a kernel is good if there is no significant loss in accuracy in the prediction error if the number of data points is halved}. This led to the introduction of 
 \[\rho=\frac{||v^{\ast}-v^{s} ||^2}{||v^{\ast} ||^2} \]
  which is the relative error between 
  $v^\ast$, the optimal recovery \eqref{eqkjelkjefffhb} of $u^\ast$ based on the full dataset
  $X=\{(x_1,y_1),\ldots,(x_N,y_N)\}$, and
  $v^s$  the optimal recovery  of both $u^\ast$ and $v^\ast$ based on half of the dataset $ X^s=\{(x_i,y_i)\mid i \in \mathcal{S}\}$ ($\operatorname{Card}(\mathcal{S})=N/2$) which admits the representation
  \begin{equation}
v^s=(y^s)^T A^s K(x^s,\cdot)
  \end{equation}
 with $y^s=\{y_i\mid i \in \mathcal{S}\}$,
 $x^s=\{x_i\mid i \in \mathcal{S}\}$,
 $A^s=(\Theta^s)^{-1}$, $\Theta^s_{i,j}=K(x_i^s,x_j^s)$.
 This quantity  $\rho$ is directly related to the game in (\ref{eqdkjdhkjhffORgameban}) where one is minimizing the relative error of $v^{\ast}$ versus $v^s$. 
Instead of using the entire the dataset $X$ one may use random subsets $X^{s_1}$ (of $X$) for $v^{\ast}$ and random subsets $ X^{s_2}$ (of $X^{s_1}$) for $v^s$. 
Writing $\sigma^2(x)=K(x,x)-K(x,X^f)K(X^f,X^f)^{-1}K(X^f,x)$ we have the pointwise error bound
\begin{equation}\label{error_estimate} |u(x)-v^\ast(x)| \leq  \sigma(x) \|u\|_{\Hc},\end{equation}
 Local error estimates such as (\ref{error_estimate}) are
classical in Kriging \cite{Wu92localerror} (see also \cite{owhadi2015bayesian}[Thm. 5.1] for applications to PDEs). $\|u\|_{\Hc}$ is bounded from below (and, in with sufficient data, can be approximated by) by $\sqrt{Y^{f,T} K(X^f,X^f)^{-1} Y^f} $, i.e., the RKHS norm of the interpolant  of $v^\ast$.

\subsection{Further numerical experiments}

\subsubsection{Changing the maximum time interval $\alpha$}

In Table \ref{tab:alphahenon}, we observe that for any $\alpha$ (the maximum time difference between observations) larger than 3, both the regular and irregular KF algorithms fail completely in the case of the Hénon map. However, the irregular KF algorithm continues to accurately predict the dynamics of the Van der Pol oscillator for $\alpha = 6$ and $7$ (Table \ref{tab:alphalorenz}), whereas the Euler and regular KF algorithms fail. In Tables \ref{tab:alphalorenz} and \ref{tab:alphalorenz8}, we see that when we vary $\alpha$ and apply the KF algorithms to the Lorenz system, while the MSE of all three versions increases as $\alpha$ is increased, the irregular KF shows a larger deterioration in performance to the extent that by $\alpha=10$, the regular KF outperforms its irregular counterpart.

\begin{table}[!tbp]
  \caption{%
   Performance of the KF algorithms for Hénon map with $\alpha=2,3$ and $4$
  }
  \label{tab:alphahenon}
  \vskip 0.1in
  \centering
  \small
  \newcommand{\NA}{---}
  \sisetup{
    detect-all           = true,
    table-format         = 2.1(2),
    separate-uncertainty = true,
    mode                 = math,
    table-text-alignment = center,
    tight-spacing,
  }
  \robustify\bfseries
  \renewrobustcmd{\bfseries}{\fontseries{b}\selectfont}
  \renewrobustcmd{\boldmath}{}
  \let\b\bfseries
  \setlength{\tabcolsep}{3.0pt}
  \begin{tabu}{lcccccc}
    \toprule
    \textsc{MAX. INTERVAL ($\alpha$)}      & \multicolumn{2}{c}{{\scriptsize\data{$\alpha$=2}}}
                         & \multicolumn{2}{c}{{\scriptsize\data{$\alpha$=3 (in Table 1)}}} & \multicolumn{2}{c}{{\scriptsize\data{$\alpha$=4}}} \\
    \midrule
     & \textsc{MSE} & \textsc{$R^2$} & \textsc{MSE} & \textsc{$R^2$} & \textsc{MSE} & \textsc{$R^2$} \\
    \midrule
     (A) \method  &  \num{0.000 \pm 0.000} & \num{0.999 \pm 0.000} & \num{0.024 \pm 0.015} & \num{0.869 \pm 0.081} &  \textsc{>>1} & \textsc{<<0}\\
          (B) KernelFlow         &  \num{0.127 \pm 0.009}         & \num{0.297 \pm 0.049}  &  \num{0.190 \pm 0.008}         & \num{-0.050 \pm 0.041} & \textsc{>>1} & \textsc{<<0}\\
    \bottomrule
  \end{tabu}
\end{table}
\begin{table}[!tbp]
  \caption{%
   Performance of KF algorithms for the Van der Pol oscillator with $\alpha=5,6$ and $7$
  }
  \label{tab:alphavdp}
  \vskip 0.1in
  \centering
  \small
  \newcommand{\NA}{---}
  \sisetup{
    detect-all           = true,
    table-format         = 2.1(2),
    separate-uncertainty = true,
    mode                 = math,
    table-text-alignment = center,
    tight-spacing,
  }
  \robustify\bfseries
  \renewrobustcmd{\bfseries}{\fontseries{b}\selectfont}
  \renewrobustcmd{\boldmath}{}
  \let\b\bfseries
  \setlength{\tabcolsep}{3.0pt}
  \begin{tabu}{lcccccc}
    \toprule
    \textsc{MAX. INTERVAL ($\alpha$)}      & \multicolumn{2}{c}{{\scriptsize\data{$\alpha$=5 (in Table 1)}}}
                         & \multicolumn{2}{c}{{\scriptsize\data{$\alpha$=6}}} & \multicolumn{2}{c}{{\scriptsize\data{$\alpha$=7}}} \\
    \midrule
     & \textsc{MSE} & \textsc{$R^2$} & \textsc{MSE} & \textsc{$R^2$} & \textsc{MSE} & \textsc{$R^2$} \\
    \midrule
     (A) \method 
    & 
         \num{0.001 \pm 0.001} & \num{0.998 \pm 0.002} & \num{0.001 \pm 0.001} & \num{0.998 \pm 0.002} & \num{0.000 \pm 0.000} & \num{0.999 \pm 0.001}\\
          (B) KernelFlow   
              &  \textsc{$>>1$} & \textsc{$<<0$} & \textsc{$>>1$} & \textsc{$<<0$} & \textsc{$>>1$} & \textsc{$<<0$} 
\\
    (C) KernelFlow (\emph{Euler}) 
                &   \textsc{$>>1$} & \textsc{$<<0$} & \textsc{$>>1$} & \textsc{$<<0$} & \textsc{$>>1$} & \textsc{$<<0$} \\
    \bottomrule
  \end{tabu}
\end{table}

\begin{table}[!tbp]
  \caption{%
   Performance of KF algorithms for  $\alpha=5,6$ and $7$, Lorenz
  }
  \label{tab:alphalorenz}
  \vskip 0.1in
  \centering
  \small
  \newcommand{\NA}{---}
  \sisetup{
    detect-all           = true,
    table-format         = 2.1(2),
    separate-uncertainty = true,
    mode                 = math,
    table-text-alignment = center,
    tight-spacing,
  }
  \robustify\bfseries
  \renewrobustcmd{\bfseries}{\fontseries{b}\selectfont}
  \renewrobustcmd{\boldmath}{}
  \let\b\bfseries
  \setlength{\tabcolsep}{3.0pt}
  \begin{tabu}{lcccccc}
    \toprule
    \textsc{MAX. INTERVAL ($\alpha$)}      & \multicolumn{2}{c}{{\scriptsize\data{$\alpha$=5 (in Table 1)}}}
                         & \multicolumn{2}{c}{{\scriptsize\data{$\alpha$=6}}} & \multicolumn{2}{c}{{\scriptsize\data{$\alpha$=7}}} \\
    \midrule
     & \textsc{MSE} & \textsc{$R^2$} & \textsc{MSE} & \textsc{$R^2$} & \textsc{MSE} & \textsc{$R^2$} \\
    \midrule
     (A) \method 
    & 
\num{0.003 \pm 0.003} & \num{0.967 \pm 0.029} & \num{0.003 \pm 0.001} & \num{0.968 \pm 0.009} & \num{0.006 \pm 0.003} & \num{0.928 \pm 0.030}  
              \\
          (B) KernelFlow   
              &  
                      \num{0.026 \pm 0.015} & \num{0.700 \pm 0.170} &
                      \num{0.017 \pm 0.017}	& \num{0.807 \pm 0.183} & \num{0.012 \pm 0.001} & \num{0.864 \pm 0.013}
\\
    (C) KernelFlow (\emph{Euler}) 
                &   \num{0.005 \pm 0.002} & \num{0.947 \pm 0.023} & \num{0.010 \pm 0.003} & \num{0.891 \pm 0.020} & \num{0.012 \pm 0.002} & \num{0.853 \pm 0.022} \\
    \bottomrule
  \end{tabu}
\end{table}

\begin{table}[!tbp]
  \caption{%
   Performance of KF algorithms for  $\alpha=8,9$ and $10$ on the Lorenz system
  }
  \label{tab:alphalorenz8}
  \vskip 0.1in
  \centering
  \small
  \newcommand{\NA}{---}
  \sisetup{
    detect-all           = true,
    table-format         = 2.1(2),
    separate-uncertainty = true,
    mode                 = math,
    table-text-alignment = center,
    tight-spacing,
  }
  \robustify\bfseries
  \renewrobustcmd{\bfseries}{\fontseries{b}\selectfont}
  \renewrobustcmd{\boldmath}{}
  \let\b\bfseries
  \setlength{\tabcolsep}{3.0pt}
  \begin{tabu}{lcccccc}
    \toprule
    \textsc{MAX. INTERVAL ($\alpha$)}      & \multicolumn{2}{c}{{\scriptsize\data{$\alpha$=8}}}
                         & \multicolumn{2}{c}{{\scriptsize\data{$\alpha$=9}}} & \multicolumn{2}{c}{{\scriptsize\data{$\alpha$=10}}} \\
    \midrule
     & \textsc{MSE} & \textsc{$R^2$} & \textsc{MSE} & \textsc{$R^2$} & \textsc{MSE} & \textsc{$R^2$} \\
    \midrule
     (A) \method 
    & 
\num{0.016 \pm 0.008} & \num{0.803 \pm 0.103} & \num{0.018 \pm 0.003} & \num{0.795 \pm 0.030} & \num{0.065 \pm 0.055} & \num{0.204 \pm 0.669}  
              \\
          (B) KernelFlow   
              &  
                      \num{0.021 \pm 0.005} & \num{0.749 +- 0.060} &
                      \num{0.024 \pm 0.003}	& \num{0.696 \pm 0.040} & \num{0.028 \pm 0.003} & \num{0.679 \pm 0.030}
\\
    (C) KernelFlow (\emph{Euler}) 
                &   \num{0.031 \pm 0.021} & \num{0.639 \pm 0.248} & \num{0.027 \pm  0.006} & \num{0.687 \pm 0.065} & \num{0.072 \pm 0.087} & \num{0.184 \pm 0.980} \\
    \bottomrule
  \end{tabu}
\end{table}

\subsubsection{\bf Computational time}

The average computational time over five repeats for the Lorenz system was 17.97 seconds for the original KF, 18.36 for the Euler KF and 21.79 for the irregular KF. Although computing the time differences increases the computational time, we deem this trade-off worthwhile given the visible improvement in accuracy seen in Figure \ref{figlorat3}.

\subsubsection{\bf Different horizons}

We re-ran the experiment for different values of the horizon $h$ while keeping $\alpha=5$. While Table \ref{tab:horizons} shows that the irregular KF algorithm remained the most accurate for $h=30$ and $h=50$, it turns out the Euler discretization is better for $h=40$.

\begin{table}[tbp]
  \caption{%
   Performance of the original Kernel Flows algorithm on the Lorenz system, the Euler discretization and the irregular Kernel Flows algorithm for horizons $h=30,40$ and $50$ and $\alpha=5$
  }
  \label{tab:horizons}
  \vskip 0.1in
  \centering
  \small
  \newcommand{\NA}{---}
  \sisetup{
    detect-all           = true,
    table-format         = 2.1(2),
    separate-uncertainty = true,
    mode                 = math,
    table-text-alignment = center,
    tight-spacing,
  }
  \robustify\bfseries
  \renewrobustcmd{\bfseries}{\fontseries{b}\selectfont}
  \renewrobustcmd{\boldmath}{}
  \let\b\bfseries
  \setlength{\tabcolsep}{3.0pt}
  \begin{tabu}{lcccccc}
    \toprule
    \textsc{HORIZON ($h$)}      & \multicolumn{2}{c}{{\scriptsize\data{$h$=30}}}
                         & \multicolumn{2}{c}{{\scriptsize\data{$h$=40}}} & \multicolumn{2}{c}{{\scriptsize\data{$h$=50}}} \\
    \midrule
     & \textsc{MSE} & \textsc{$R^2$} & \textsc{MSE} & \textsc{$R^2$} & \textsc{MSE} & \textsc{$R^2$} \\
    \midrule
     (A) \method 
    & 
         \num{0.003 \pm 0.001}         &  \num{0.962 \pm 0.013}  & \num{0.050 \pm 0.058} & \num{0.429 \pm 0.666} & \num{0.018 \pm 0.006} & \num{0.791 \pm 0.074}\\
          (B) KernelFlow   
              &  \num{0.015 \pm 0.007} & \num{0.826 \pm 0.087} & \num{0.037 \pm 0.023} & \num{0.571 \pm 0.266} & \num{0.023 \pm 0.004} & \num{0.734 \pm 0.031}
\\
    (C) KernelFlow (\emph{Euler}) 
                &  \num{0.012 \pm 0.001}         & \num{0.871 \pm 0.014}  & \num{0.015 \pm 0.000} & \num{0.826 \pm 0.005} & \num{0.025 \pm 0.000} & \num{0.717 \pm 0.000}\\
    \bottomrule
  \end{tabu}
\end{table}

\subsubsection{Varying the learning rate}

\begin{table}[tbp]
  \caption{%
   Performance of the irregular Kernel Flows algorithm on the Lorenz system for learning rate $\eta=0.001,0.01$ and $0.1$
where $h=20$ and $\alpha=5$   }
  \label{tab:lr}
  \vskip 0.1in
  \centering
  \small
  \newcommand{\NA}{---}
  \sisetup{
    detect-all           = true,
    table-format         = 2.1(2),
    separate-uncertainty = true,
    mode                 = math,
    table-text-alignment = center,
    tight-spacing,
  }
  \robustify\bfseries
  \renewrobustcmd{\bfseries}{\fontseries{b}\selectfont}
  \renewrobustcmd{\boldmath}{}
  \let\b\bfseries
  \setlength{\tabcolsep}{3.0pt}
  \begin{tabu}{lcccccc}
    \toprule
    \textsc{LEARNING RATE ($\eta$)}      & \multicolumn{2}{c}{{\scriptsize\data{$\eta$=0.001}}}
                         & \multicolumn{2}{c}{{\scriptsize\data{$\eta$=0.01 (in Table 1)}}} & \multicolumn{2}{c}{{\scriptsize\data{$\eta$=0.1}}} \\
    \midrule
     & \textsc{MSE} & \textsc{$R^2$} & \textsc{MSE} & \textsc{$R^2$} & \textsc{MSE} & \textsc{$R^2$} \\
    \midrule
     (A) \method 
    & 
         \num{0.035 \pm 0.033}           & \num{0.606 \pm 0.374} & \num{0.003 \pm 0.003}        & \num{0.967 \pm 0.029} & \num{0.080 \pm 0.155} & \num{0.606 \pm 0.374}\\
    \bottomrule
  \end{tabu}
\end{table}

When we varied the learning rate $\eta$ from $0.01$ to $0.001$ for the irregular KF algorithm on the Lorenz system while keeping the number of iterations at 1000, the MSE increased as irregular KF did not sufficiently learn from the training data in that period (see Table \ref{tab:lr}). When $\eta$ was increased to $0.1$, the performance was also worse compared to our default $\eta=0.01$ as the algorithm overshot the minimum. Therefore, we can conclude that $\eta=0.01$ is a sensible learning rate for training for 1000 epochs.

\subsubsection{Varying the number of training epochs}

  %

\begin{figure}[!htb]
     \centering
     \includegraphics[width=.95\linewidth]{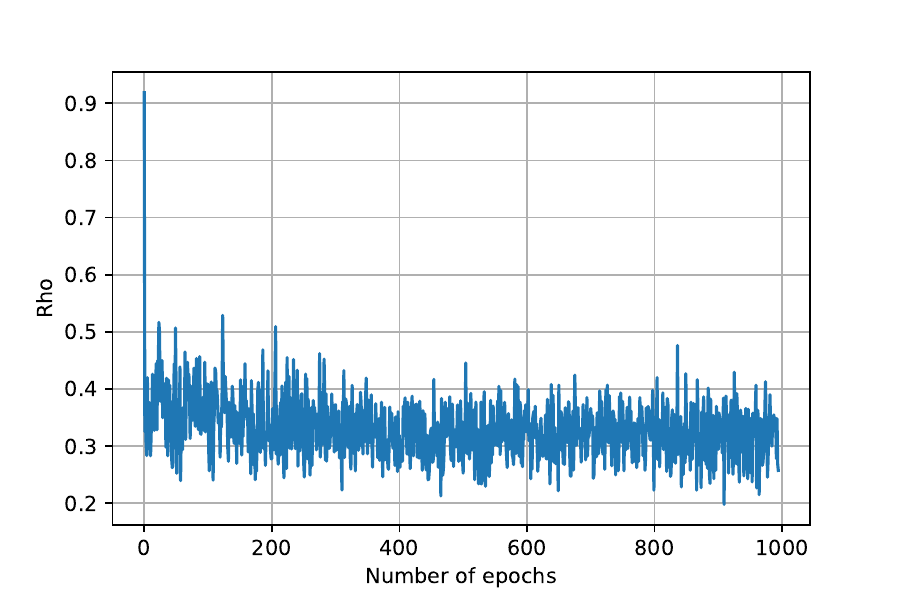}
     \caption{Lorenz system, evolution of $\rho(\theta)$ against the number of training epochs}\label{Fig:Data11}
\end{figure}

Figure \ref{Fig:Data11} shows that when we changed the number of training epochs, $\rho(\theta)$ does not consistently decrease against the number of training epochs after an initial drop. This is because the irregular KF algorithm has already learned the kernel within the first few epochs, and the fluctuations of $\rho(\theta)$ in \ref{Fig:Data11} are due to the inherent randomness of the mini-batch stochastic gradient descent (SGD) optimizer that was used in training.

\subsection{Acknowledgments}
HO and BH gratefully acknowledges partial support by the Air Force Office of Scientific Research under MURI award number FA9550-20-1-0358 (Machine Learning and Physics-Based Modeling and Simulation) and JPL/NASA (Greenland contribution to sea level by 2050: The role of meltwater in shaping the future ice sheet evolution, UQ-aware Machine Learning for Uncertainty Quantification). EDB is funded by a FWO-SB grant. 

\subsection{Code}

All the relevant code for the experiments can be found at:
\\
https://github.com/jlee1998/Kernel-Flows-for-Irregular-Time-Series
\bibliography{sample.bib}

\bibliographystyle{plain}

\end{document}